\documentclass[final,5p,times,twocolumn]{elsarticle}

\usepackage{prletter}
\usepackage{amsmath,amssymb,amsthm}
\usepackage{graphicx}
\usepackage{booktabs}
\usepackage{url}
\usepackage{hyperref}
\usepackage{tikz}
\usetikzlibrary{arrows.meta,positioning,calc,shapes.geometric,fit,backgrounds,decorations.pathmorphing}

\makeatletter
    {\def\@noitemerr{\@latex@warning{Empty `thebibliography' environment}}%
     \endlist}
\makeatother

\newcommand{\firstclaim}{to the best of our knowledge, the first peer-reviewed reinforcement learning trading system whose state representation is built explicitly from Steidlmayer--Dalton Auction Market Theory primitives}

\journal{arXix}

\begin{document}

%% =====================================================================
%% FRONTMATTER
%% =====================================================================

\begin{frontmatter}

\title{ViperQ: Order Flow Pattern Recognition via\\ Auction Market Theory for Reinforcement Learning Trading}

\author[uccs]{Asser Moustafa\corref{cor1}}
\ead{amoustaf@uccs.edu}

\author[uiowa]{Rares-Mihail Neagu}
\ead{rneagu@uiowa.edu}

\author[uccs]{Jugal Kalita}
\ead{jkalita@uccs.edu}

\cortext[cor1]{Corresponding author.}

\address[uccs]{Department of Computer Science, University of Colorado, Colorado Springs, 1420 Austin Bluffs Pkwy, Colorado Springs, CO 80918, USA}
\address[uiowa]{Department of Computer Science, University of Iowa, 101 Jessup Hall, Iowa City, IA 52242, USA}

\begin{abstract}
Reinforcement learning trading systems published in the academic literature overwhelmingly rely on price-aggregate state representations (OHLCV bars) or limit-order-book depth features, leaving microstructure pattern theories from the practitioner literature, namely Auction Market Theory and Market Profile, without a peer-reviewed computational instantiation. We present \textbf{ViperQ}, a reinforcement learning system whose state representation is built explicitly from Auction Market Theory primitives: Volume Point of Control, Value Area position, Low Volume Node flags, Cumulative Volume Delta divergence, and tape-velocity signatures, assembled into a 20-dimensional Z-normalised vector. Two Proximal Policy Optimisation agents are trained with a prospect-theory-grounded asymmetric reward function that penalises losing holds at a magnitude consistent with Kahneman and Tversky's loss-aversion coefficient. Evaluated on a held-out twelve-month partition of institutional tick data the agents have never seen, ViperQ achieves $+163.6\%$ ROI on TSLA ($-27.5\%$ max drawdown, 27{,}019 trades) and $+116.5\%$ ROI on NVDA ($-47.8\%$ max drawdown, 12{,}892 trades) under zero leverage. The results establish Auction Market Theory features as a tractable structured input modality for sequential decision-making on financial time series and motivate further work on microstructure-aware policy learning.
\end{abstract}


\end{frontmatter}

%% =====================================================================
%% 1. INTRODUCTION
%% =====================================================================
\section{Introduction}
\label{sec:intro}

The application of deep reinforcement learning (RL) to financial decision-making has expanded rapidly over the past five years~\cite{hambly2023recent, pippas2025evolution}, with documented systems spanning portfolio rebalancing~\cite{yang2020ensemble, liu2020finrl}, optimal trade execution~\cite{lin2020endtoend, hafsi2024optimal, espana2025queue}, market making~\cite{wang2024marketmaking, zimmer2025marketmaking}, and directional high-frequency trading~\cite{briola2021drlhft, zhang2020deeprl}. Across this body of work, the choice of state representation has been remarkably homogeneous. Systems either consume price-aggregate features (open-high-low-close bars, returns, volatility, technical indicators) or, when operating at higher resolution, consume raw limit-order-book (LOB) depth snapshots~\cite{zhang2019deeplob, xiao2025lit}. A parallel tradition reads the same data through the lens of \emph{market microstructure theory}: price discovery is the outcome of strategic interaction between informed and uninformed order flow~\cite{kyle1985continuous}, market makers update quotes to reflect the information content of incoming orders~\cite{glosten1985bid}, and the resulting microstructure dynamics determine when and how prices transition between equilibria~\cite{ohara1995market, easley2012flow}. Within this microstructure framing, Steidlmayer's Market Profile~\cite{steidlmayer1986markets} and Dalton's Auction Market Theory~\cite{dalton1990mind} (AMT) operationalise two repeatable microstructure regimes that have been widely adopted in proprietary-trading practice: \emph{balance} states (mean-reverting consensus on fair value) and \emph{imbalance} states (one-sided liquidity vacuums driving price discovery to a new equilibrium). Despite three-and-a-half decades of practitioner adoption, this microstructure pattern theory has not, so far as exhaustive search of arXiv, Semantic Scholar, IEEE Xplore, ACM Digital Library, and ScienceDirect can determine, been instantiated as a state representation for a peer-reviewed reinforcement learning trading agent.

This paper closes that gap. We present \textbf{ViperQ}, \firstclaim. The state representation encodes Volume Point of Control (VPOC), Value Area boundaries, Low Volume Node flags, Cumulative Volume Delta divergence, tape velocity, and order-flow imbalance as a 20-dimensional Z-normalised vector that is consumed directly by a Proximal Policy Optimisation~\cite{schulman2017ppo} agent operating at 1-second resolution on institutional tick data.

The framing of the problem is, deliberately, a pattern recognition framing rather than a financial-engineering framing. The classifier-equivalent task is the binary recognition of \emph{balance versus imbalance} regimes from a stream of microstructure observations, with sequential policy learning conditional on that recognition. The contribution we believe to be of broadest interest to the pattern recognition community is therefore the structured state representation itself: a demonstration that a body of practitioner pattern theory, encoded explicitly rather than learned end-to-end, yields a state space on which a small neural policy can converge to substantial out-of-sample profit. The trading-domain application is, in this view, downstream of the methodological point.

We summarise our contributions as follows.

\begin{itemize}
\item \textbf{Auction-Market-Theory state representation.} We define and operationalise a 20-dimensional state vector $s_t \in \mathbb{R}^{20}$ whose physics-derived components (VPOC, Value Area position, Low Volume Node flags, CVD divergence, tape velocity, order-flow imbalance) are direct computational analogues of constructs first formalised by Steidlmayer~\cite{steidlmayer1986markets} and Dalton et al.~\cite{dalton1990mind}. To the best of our knowledge, no prior peer-reviewed RL trading agent has used this feature family as state input.
\item \textbf{Prospect-theory-grounded reward shaping.} The Sniper reward function applies asymmetric pain penalties consistent with Kahneman and Tversky's loss-aversion coefficient~\cite{kahneman1979prospect, tversky1992advances}, augmented by a synthetic in-training transaction-fee inflation that forces the agent toward high-conviction setups rather than micro-scalping.
\item \textbf{Asset-specific behavioural divergence.} We show that a single architecture, trained with two reward-function configurations, converges to two distinct profitable policies (a high-frequency scalper for TSLA and a trend-following risk-averse agent for NVDA), demonstrating that asset-specific reward shaping yields different and individually profitable local optima from the same underlying state representation.
\item \textbf{Out-of-sample empirical validation.} On a strictly held-out 12-month test partition, ViperQ produces $+163.6\%$ ROI on TSLA and $+116.5\%$ ROI on NVDA with zero leverage and regulatory-fee accounting. These figures lie above the typical range surveyed by Pippas et al.~\cite{pippas2025evolution}, at the cost of materially deeper drawdowns than conservative portfolio-rebalancing systems (Section~\ref{sec:results}).
\end{itemize}

The remainder of the paper is organised as follows. Section~\ref{sec:related} situates ViperQ against the deep-learning-on-LOB and RL-trading literature. Section~\ref{sec:method} formalises the state representation, the reward function, and the dual-PPO training strategy. Section~\ref{sec:risk} describes the risk-bounded execution layer. Section~\ref{sec:setup} details the experimental protocol; Section~\ref{sec:results} presents results and the comparison with prior systems. Section~\ref{sec:limitations} states the limitations of the present work, and Section~\ref{sec:conclusion} concludes.

%% =====================================================================
%% 2. RELATED WORK
%% =====================================================================
\section{Related Work}
\label{sec:related}

\textbf{Deep learning on limit order books.} The DeepLOB architecture~\cite{zhang2019deeplob} established convolutional networks as the canonical baseline for mid-price prediction from raw LOB snapshots, and has been extended to transformer architectures (LiT~\cite{xiao2025lit}) and to multi-asset attention models. These systems treat the order book as a structured but unannotated time series and learn features end-to-end; they do not encode prior microstructure theory and they typically operate as classifiers rather than as control agents.

\textbf{Reinforcement learning for trading and execution.} A second strand applies RL directly to trading or execution decisions. Zhang et al.~\cite{zhang2020deeprl} compared DQN, policy-gradient, and A2C agents on 50 futures contracts at daily resolution. Th\'eate and Ernst~\cite{theate2021application} introduced TDQN for daily equity trading. Yang et al.~\cite{yang2020ensemble} demonstrated an ensemble of PPO, A2C and DDPG on the Dow 30 with daily rebalancing. Liu et al.~\cite{liu2020finrl} released the FinRL library, which standardised much of the subsequent literature. At higher frequency, Briola et al.~\cite{briola2021drlhft} trained PPO on Intel LOB events; Lin and Beling~\cite{lin2020endtoend} produced an end-to-end PPO execution agent on Level-2 data; Hafsi and Vittori~\cite{hafsi2024optimal} and Espana et al.~\cite{espana2025queue} continue this line into queue-reactive simulation environments. Coletta et al.~\cite{coletta2023conditional} provided conditional generators for LOB environments; Wang et al.~\cite{wang2024marketmaking,mascioli2024pymarketsim,zimmer2025marketmaking} addressed RL market making. Two contemporary surveys~\cite{hambly2023recent, pippas2025evolution} catalogue the field; Pippas et al.\ critically evaluate 167 publications and identify microstructure-feature integration as a methodological gap.

\textbf{Microstructure-feature work.} Jaddu and Bilokon~\cite{jaddu2024deep} combine deep-learning representations of order books with RL using Order Flow Imbalance features in the sense of~\cite{cont2014price}. To our knowledge this is the closest prior work to ViperQ in spirit, but the feature family is OFI rather than Auction Market Theory; the constructs we use (VPOC, Value Area, LVN, balance/imbalance classification) do not appear in any reviewed system.

\textbf{Auction Market Theory.} The theoretical tradition we draw on originates with Steidlmayer's Market Profile~\cite{steidlmayer1986markets} and Dalton's Auction Market Theory~\cite{dalton1990mind}, which frame intraday price discovery as oscillation between balance (mean-reverting two-way trade) and imbalance (one-sided liquidity vacuum) states. These constructs have been extensively adopted in proprietary-trading practice for over three decades, but a focused literature search across the major scholarly indexes returned no peer-reviewed reinforcement-learning system that uses them as explicit state features. ViperQ closes that gap.

\textbf{Pattern recognition precedents.} Application-driven reinforcement learning has prior precedent in this journal. Liu et al.~\cite{liu2024hierarchical} apply hierarchical RL to chip-macro placement; Tsantekidis et al.~\cite{tsantekidis2021diversity} apply knowledge-distilled deep RL to financial trading in a related Elsevier venue. These works establish that domain-application RL contributions are within scope when framed as pattern recognition methodology.

%% =====================================================================
%% 3. METHOD
%% =====================================================================
\section{Method}
\label{sec:method}

We treat sequential trading as a Markov Decision Process $(\mathcal{S}, \mathcal{A}, P, R, \gamma)$ implemented as a custom Gymnasium~\cite{towers2024gymnasium} environment,\footnote{Gymnasium: \url{https://gymnasium.farama.org/}} with all execution priced at the next bar's open to eliminate look-ahead bias.

\subsection{Order flow patterns: matching engine and Auction Market Theory}
\label{sec:method_patterns}

\begin{figure}[t]
\centering
\includegraphics[width=\columnwidth]{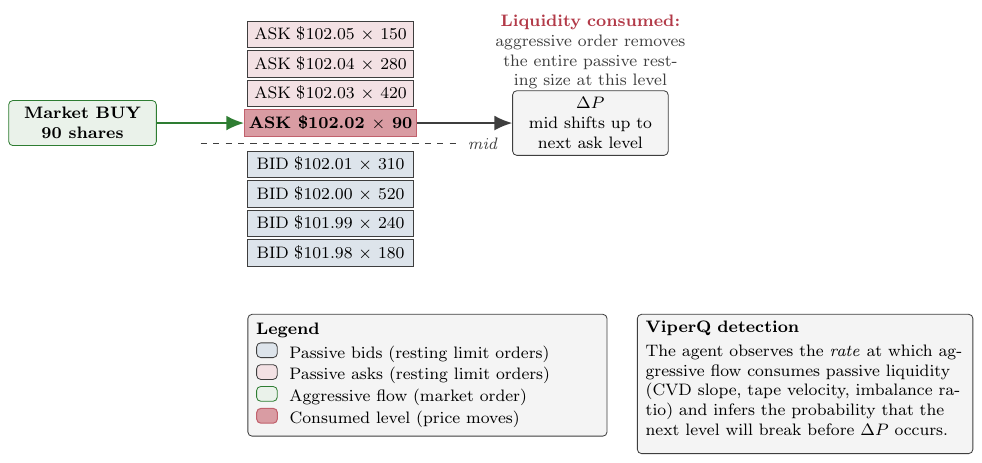}
\caption{Matching engine: aggressive market flow consumes passive resting liquidity; price moves only when the resting size at the touch is fully consumed. ViperQ observes the consumption rate via CVD slope, tape velocity, and imbalance ratio (Section~\ref{sec:method_state}).}
\label{fig:matching}
\end{figure}

The state representation rests on two compositional patterns from market microstructure theory. Figure~\ref{fig:matching} formalises the matching engine: a continuous double auction~\cite{kyle1985continuous, gould2013limit} in which aggressive market orders consume passive resting liquidity, with mid-price displacement $\Delta P$ occurring only when the resting size at the touch is exhausted~\cite{cont2014price}. ViperQ measures aggression and passive depth directly, via the Cumulative Volume Delta, tape velocity, and order-flow imbalance features defined below, and so observes the consumption rate \emph{before} $\Delta P$ becomes visible in price.

\begin{figure}[t]
\centering
\includegraphics[width=\columnwidth]{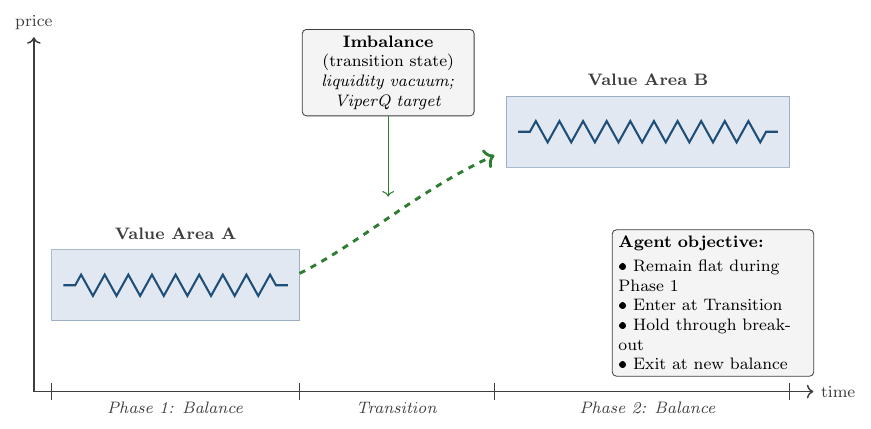}
\caption{Auction Market Theory: the market oscillates between balance and imbalance. The agent targets the transition from Value Area A to Value Area B, remaining flat during balance phases and holding through breakouts.}
\label{fig:amt}
\end{figure}

Figure~\ref{fig:amt} depicts Auction Market Theory's second pattern: intraday price discovery oscillates between balance regimes (mean-reverting trade inside a value area; passive policy optimal: remain flat) and imbalance regimes (one-sided aggression creates a liquidity vacuum and price sprints to find a new value area)~\cite{steidlmayer1986markets, dalton1990mind}. The agent's objective is detecting the transition state.

\subsection{State representation}
\label{sec:method_state}

Each observation $s_t \in \mathbb{R}^{20}$ is a Z-score-normalised vector grouped into three blocks (Table~\ref{tab:state}). Block A encodes the AMT/microstructure patterns; Block B injects two dynamic position-aware variables at runtime; Block C provides nine lagged log-returns to give a 9-second memory without recurrence.

\begin{table}[t]
\centering
\small
\caption{The 20-dimensional state vector $s_t$. Block A encodes Auction Market Theory and microstructure patterns; Block B is injected at runtime from the agent's position; Block C supplies short-horizon memory.}
\label{tab:state}
\begin{tabular}{@{}clp{4.4cm}@{}}
\toprule
\textbf{Dim} & \textbf{Name} & \textbf{Construct} \\
\midrule
\multicolumn{3}{l}{\textit{Block A: AMT / microstructure patterns}} \\
1 & \texttt{z\_price\_vwap}    & Z(price$-$VWAP), gravity to fair value \\
2 & \texttt{z\_price\_vpoc}    & Z(price$-$VPOC), distance to Volume Point of Control \\
3 & \texttt{dist\_to\_wall}    & Normalised position within 300\,s high--low range \\
4 & \texttt{cvd\_slope}        & 60\,s first-difference of CVD \\
5 & \texttt{cvd\_divergence}   & Binary flag on 300\,s price--CVD correlation $< -0.5$ \\
6 & \texttt{tape\_velocity}    & Trade count $\div$ 60\,s rolling mean \\
7 & \texttt{trade\_size\_z}    & Z-score of volume over 300\,s window \\
8 & \texttt{imbalance\_ratio}  & $\delta / (V + \varepsilon)$, buy vs.\ sell dominance \\
9 & \texttt{in\_lvn\_zone}     & Binary flag for Low Volume Node ($V < 0.5 \bar V$) \\
\midrule
\multicolumn{3}{l}{\textit{Block B: dynamic position state}} \\
10 & \texttt{unrealized\_pnl}  & $(P_t - P_{\text{entry}}) \cdot \text{pos}$ \\
11 & \texttt{time\_in\_trade}  & Seconds since entry \\
\midrule
\multicolumn{3}{l}{\textit{Block C: lagged log-returns}} \\
12--20 & \texttt{lag\_1\,\ldots\,lag\_9} & $\log(P_t / P_{t-k})$, $k \in \{1,\ldots,9\}$ \\
\bottomrule
\end{tabular}
\end{table}

VPOC is approximated by a one-hour rolling VWAP, following standard practitioner usage; VWAP itself follows~\cite{cartea2015algorithmic}. Z-score normalisation over rolling windows ensures stationarity across volatility regimes.

\subsection{Sniper reward function}
\label{sec:method_reward}

The action space is $\mathcal{A} = \{\text{flat},\,\text{long}\}$; short positions are disabled. The reward $R_t$ aggregates six components.

\noindent\textbf{(1) Realised PnL at trade close.}
\begin{equation}
R_{\text{close}} = \alpha \cdot (P_{\text{exit}} - P_{\text{entry}}) \cdot n \cdot m - \alpha \cdot f_{\text{syn}} \cdot n,
\label{eq:close}
\end{equation}
where $\alpha = 0.01$, $n$ is share count, $m$ the contract multiplier, and $f_{\text{syn}} = \$0.50$/share is a synthetic in-training transaction fee approximately $2{,}500\times$ the real regulatory fee.

\noindent\textbf{(2) Asymmetric unrealised-PnL shaping.}
\begin{equation}
R_{\text{hold}} = \begin{cases}
+0.001 \cdot U_t, & U_t > 0 \\
-0.05 - 0.005 \cdot |U_t|, & U_t \leq 0
\end{cases}
\label{eq:hold}
\end{equation}
The asymmetry between the two branches is grounded in Prospect Theory~\cite{kahneman1979prospect, tversky1992advances}: Tversky and Kahneman's empirical estimate of the loss-aversion coefficient $\lambda \approx 2.25$ (losses weighted approximately $2.25\times$ more heavily than equivalent gains) provides the theoretical basis for asymmetric reward shaping. The effective asymmetry in Eq.~\eqref{eq:hold} is stronger than the $\lambda \approx 2.25$ baseline, deliberately so, to compensate for the additional reward density produced by 1-second decision steps and to suppress the disposition effect~\cite{lin2020endtoend, hafsi2024optimal}.

\noindent\textbf{(3) Trend bonus.} When $U_t > \$100$, $R_{\text{trend}} = +0.10$ per step, encouraging the agent to hold through sustained momentum rather than taking premature profits.

\noindent\textbf{(4) Time decay.} After 3600\,s, $R_{\text{time}} = -0.001 \cdot (t_{\text{hold}} - 3600)$ penalises stagnant positions.

\noindent\textbf{(5) Stop loss.} $R_{\text{stop}} = -50.0$ if $U_t < -\$200$, triggering forced exit.

\noindent\textbf{(6) Drawdown shock and termination.} During training, when portfolio drawdown $< -2\%$, a continuous $-2.0$/step penalty is applied. If drawdown breaches the episode limit, $R_{\text{dd}} = -100.0$ and the episode terminates.

\subsection{Asset-specific dual-PPO training}
\label{sec:method_dual}

Two PPO agents~\cite{schulman2017ppo} are trained from the same state representation but with two reward configurations (Table~\ref{tab:configs}). The TSLA agent operates with the default trend bonus and drawdown shock; the NVDA agent receives an additional $+0.10$/step held-winner bonus and the continuous drawdown shock at the same threshold, producing a more risk-averse trend-following policy. Both share the 300-second minimum holding constraint. PPO defaults follow Stable-Baselines3 implementation of~\cite{schulman2017ppo}.

\textbf{Why PPO.} The algorithm was selected over alternatives on four grounds. (i) The action space is discrete ($\{\text{flat},\,\text{long}\}$), which rules out continuous-control methods such as DDPG (used inside the Yang et al.~\cite{yang2020ensemble} ensemble) and SAC without an additional discretisation layer. (ii) DQN-family methods including TDQN~\cite{theate2021application} rely on an off-policy replay buffer that mixes transitions across non-stationary market regimes; PPO's on-policy clipped updates avoid this distributional-shift bias. (iii) A2C exhibits higher gradient variance and is less sample-efficient at the data volumes used here (49.9\,M test bars across both assets and roughly four times that in training). (iv) PPO is the dominant choice in published microstructure-RL precedents (Lin and Beling~\cite{lin2020endtoend}, Briola et al.~\cite{briola2021drlhft}, Wang et al.~\cite{wang2024marketmaking}), supporting comparability with prior work. All numerical constants in Eqs.~\eqref{eq:close}--\eqref{eq:hold} and items (3)--(6) above, together with the per-asset configurations in Table~\ref{tab:configs}, are summarised in Table~\ref{tab:hyperparams} with their selection mechanism (validation slice, industry convention, theory-derived, or library default).

\begin{table}[t]
\centering
\small
\caption{Asset-specific training configurations. The same architecture and state vector yield two distinct profitable policies. All constants listed here, with their selection mechanism, appear in Table~\ref{tab:hyperparams}.}
\label{tab:configs}
\begin{tabular}{@{}lll@{}}
\toprule
\textbf{Constraint}     & \textbf{TSLA agent}      & \textbf{NVDA agent}      \\
\midrule
Min.\ hold (s)          & 300                      & 300                      \\
Drawdown shock          & $-2.0$/step              & $-2.0$/step              \\
Trend bonus             & none                     & $+0.10$/step ($U_t > \$100$) \\
\bottomrule
\end{tabular}
\end{table}

\begin{figure}[t]
\centering
\includegraphics[width=0.85\columnwidth]{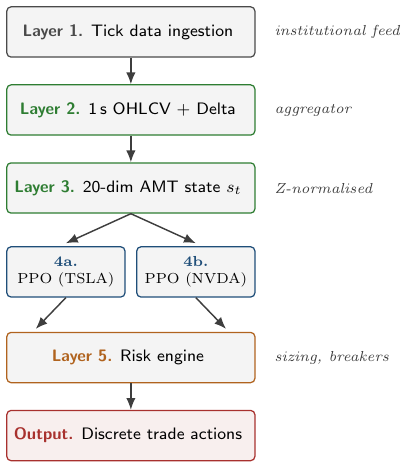}
\caption{ViperQ system architecture: pattern extraction (Layers 1--3), policy inference (Layer 4, dual PPO), and risk-bounded execution (Layer 5 and output).}
\label{fig:arch}
\end{figure}

The complete 5-layer pipeline is summarised in Figure~\ref{fig:arch}: ingestion of institutional tick data, 1-second OHLCV aggregation with order-flow metrics, the 20-dimensional state engineering described above, dual PPO inference, and a risk-bounded execution layer detailed in Section~\ref{sec:risk}.

%% =====================================================================
%% 4. RISK-BOUNDED EXECUTION
%% =====================================================================
\section{Risk-bounded execution}
\label{sec:risk}

The risk layer translates the agent's discrete action into a final share count and enforces three circuit breakers. Position size follows volatility-adjusted sizing:
\begin{equation}
n_{\text{ideal}} = \left\lfloor \frac{B_t \cdot r_a}{2 \cdot \text{ATR}_t} \right\rfloor,
\quad n_{\text{final}} = \min\!\left(n_{\text{ideal}},\, \left\lfloor C_t / P_t \right\rfloor\right),
\label{eq:sizing}
\end{equation}
where $B_t$ is current equity, $r_a \in \{0.25\%, 0.50\%, 1.0\%\}$ maps conviction to dollar-risk, ATR$_t$ is the prevailing Average True Range, and $C_t$ is available cash. The cash clamp in the second term makes leverage impossible. Three additional hard rules over-ride the agent: (i) daily loss limit of $-2\%$ of starting equity locks trading until the next session; (ii) per-trade unrealised loss exceeding $\$200$ triggers forced exit; (iii) all positions are flattened at 15:55 ET to eliminate overnight gap risk. The $-2\%$ daily loss limit, the 2-ATR stop distance, and the end-of-day flatten are standard intraday proprietary-trading conventions rather than hand-tuned parameters.

%% =====================================================================
%% 5. EXPERIMENTAL SETUP
%% =====================================================================
\section{Experimental setup}
\label{sec:setup}

\textbf{Data.} Institutional tick data for NVDA and TSLA from Databento, January 2020 to February 2026, covering the COVID-era volatility regime, the 2023--24 AI-driven rally, and the 2025--26 rate-cycle period. Nanosecond ticks are aggregated to 1-second OHLCV bars augmented with Delta (signed aggressive volume) and trade count. An 80/20 chronological partition holds out the final twelve months; no information from the test partition enters training or hyperparameter selection.

\textbf{Hyperparameter selection.} Every numerical constant introduced in Sections~\ref{sec:method_reward}--\ref{sec:risk} is treated as a hyperparameter and listed in Table~\ref{tab:hyperparams} with the mechanism by which its value was fixed. Four mechanisms are used: \textbf{VAL} (selected on the validation portion of the training split, in the spirit of~\cite{lin2020endtoend, briola2021drlhft, zhang2020deeprl, yang2020ensemble}); \textbf{CONV} (industry convention for intraday proprietary trading, taken without tuning); \textbf{TH} (theory-derived, with a literature anchor); \textbf{LIT} (library default, used as-is from Stable-Baselines3 for the PPO architecture, following~\cite{schulman2017ppo}). A formal one-factor-at-a-time sensitivity sweep is reserved for future work, in line with Lin and Beling~\cite{lin2020endtoend}, who in a comparable setting note that they ``did not perform an exhaustive grid search on the hyperparameter space, but rather [drew] random samples from the hyperparameter space due to limited computing resources''.

\begin{table}[t]
\centering
\scriptsize
\caption{Master hyperparameter table. \textbf{Mech.}: \emph{VAL} = selected on the validation slice of the training partition; \emph{CONV} = industry convention for intraday proprietary trading, taken without tuning; \emph{TH} = theory-derived with literature anchor; \emph{LIT} = library default (Stable-Baselines3).}
\label{tab:hyperparams}
\setlength{\tabcolsep}{3pt}
\renewcommand{\arraystretch}{0.9}
\begin{tabular}{@{}p{2.7cm}lcl@{}}
\toprule
\textbf{Element} & \textbf{Value} & \textbf{Mech.} & \textbf{Source / rationale} \\
\midrule
\multicolumn{4}{l}{\textit{Reward function (Eqs.~\ref{eq:close}--\ref{eq:hold}, items (3)--(6))}} \\
$\alpha$ (PnL scaling)         & $0.01$              & VAL  & PPO gradient stability \\
$f_{\text{syn}}$               & \$0.50/share        & VAL  & suppresses sub-tick scalping (\S\ref{sec:results_behaviour}) \\
Hold gain branch               & $+0.001\cdot U_t$   & VAL  & disposition-effect tuning \\
Hold loss branch               & $-0.05 - 0.005|U_t|$ & TH   & Tversky--Kahneman $\lambda{\approx}2.25$~\cite{tversky1992advances} \\
Trend bonus                    & $+0.10$/step        & VAL  & held-winner reinforcement \\
Trend threshold                & $\$100$ ($U_t$)     & CONV & $1\%$ of $\$10$k capital \\
Time-decay rate                & $-0.001$/step       & VAL  & validation \\
Time-decay onset               & $3600$\,s           & CONV & 1\,h intraday cap \\
Stop-loss penalty              & $-50.0$             & VAL  & validation \\
Stop-loss threshold            & $-\$200$ ($U_t$)    & CONV & $2\%$ of capital \\
Drawdown shock                 & $-2.0$/step         & VAL  & validation \\
Drawdown threshold             & $-2\%$              & CONV & prop daily-loss limit \\
Episode-end penalty            & $-100.0$            & VAL  & validation \\
\midrule
\multicolumn{4}{l}{\textit{Risk-bounded execution (\S\ref{sec:risk}, Eq.~\ref{eq:sizing})}} \\
Per-trade risk $r_a$           & $\{0.25, 0.50, 1.0\}\%$ & CONV & conviction sizing \\
ATR stop multiplier            & $2{\times}$         & CONV & 2-ATR distance \\
Daily loss cap                 & $-2\%$              & CONV & prop daily limit \\
EOD flatten                    & 15:55 ET            & CONV & gap-risk hedge \\
\midrule
\multicolumn{4}{l}{\textit{State construction (Table~\ref{tab:state})}} \\
VWAP/VPOC window               & $3600$\,s           & CONV & practitioner default \\
Microstructure window          & $300$\,s            & CONV & 5-min standard \\
CVD / velocity window          & $60$\,s             & CONV & 1-min standard \\
Lag depth                      & 9 steps             & VAL  & memory/dim.\ tradeoff \\
\midrule
\multicolumn{4}{l}{\textit{Training}} \\
PPO architecture               & SB3 defaults        & LIT  & Stable-Baselines3~\cite{schulman2017ppo} \\
Min.\ hold (training)          & $300$\,s            & CONV & 5-min HFT/momentum cut \\
\bottomrule
\end{tabular}
\end{table}

\textbf{Evaluation.} Initial capital is \$10{,}000 per asset. Backtests are deterministic; all trades clear at the next bar's open. Real regulatory fees ($\$0.0002$/share) are charged at evaluation time. The synthetic fee $f_{\text{syn}}$ used during training (Eq.~\ref{eq:close}) is \emph{not} active at evaluation.

%% =====================================================================
%% 6. RESULTS
%% =====================================================================
\section{Results}
\label{sec:results}

\subsection{Out-of-sample performance}
Table~\ref{tab:results} reports performance on the held-out twelve-month partition. The TSLA agent achieves $+163.6\%$ ROI with $-27.5\%$ maximum drawdown across 27{,}019 trades; the NVDA agent achieves $+116.5\%$ ROI with $-47.8\%$ maximum drawdown across 12{,}892 trades. Both are produced with zero leverage and full regulatory-fee accounting.

\begin{table}[t]
\centering
\small
\caption{ViperQ performance on the held-out test partition (final 20\% of data, never seen during training). Initial capital \$10{,}000 per asset; zero leverage; \$0.0002/share regulatory fees applied.}
\label{tab:results}
\begin{tabular}{@{}lrr@{}}
\toprule
\textbf{Metric}            & \textbf{TSLA}    & \textbf{NVDA}    \\
\midrule
Final balance              & \$26{,}364.87    & \$21{,}652.86    \\
PnL                        & $+$\$16{,}364.87 & $+$\$11{,}652.86 \\
ROI                        & $+163.6\%$       & $+116.5\%$       \\
Maximum drawdown           & $-27.5\%$        & $-47.8\%$        \\
Total trades               & 27{,}019         & 12{,}892         \\
Test-set bars              & 33.4\,M          & 16.5\,M          \\
\bottomrule
\end{tabular}
\end{table}

\subsection{Comparison with published RL trading systems}
\label{sec:results_comparison}

\begin{figure}[t]
\centering
\includegraphics[width=\columnwidth]{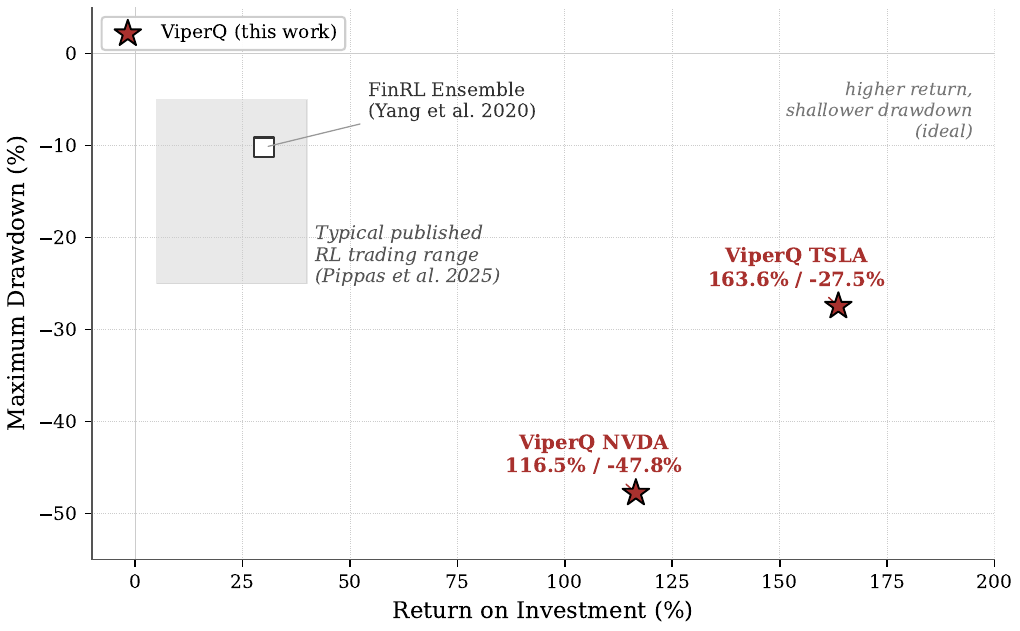}
\caption{ViperQ in the ROI/MDD plane against published RL trading systems with concrete numbers. The shaded region marks the typical published range surveyed in~\cite{pippas2025evolution}.}
\label{fig:scatter}
\end{figure}

Direct head-to-head comparison is precluded by the absence of standardised HFT-RL benchmarks: published papers differ in assets, time horizons, data resolution, capital assumptions, and reporting conventions. Pippas et al.~\cite{pippas2025evolution} survey 167 RL-finance papers and observe that consistent benchmarking is itself an open methodological problem. We therefore position ViperQ within the published landscape rather than claim dominance over it. Table~\ref{tab:literature} lists peer-comparable systems with retrievable performance numbers; Figure~\ref{fig:scatter} plots ViperQ alongside those points in the ROI/MDD plane.

\begin{table}[t]
\centering
\scriptsize
\caption{ViperQ vs.\ recently published RL trading systems with retrievable performance numbers. Typical published range~\cite{pippas2025evolution}: ROI $[5\%, 40\%]$, MDD $[-5\%, -25\%]$. ``n.r.''$=$not reported; ``n.a.''$=$not applicable.}
\label{tab:literature}
\setlength{\tabcolsep}{3pt}
\renewcommand{\arraystretch}{1.05}
\begin{tabular}{@{}lllrr@{}}
\toprule
\textbf{System} & \textbf{Algorithm} & \textbf{Asset / Horizon} & \textbf{ROI \%} & \textbf{MDD \%} \\
\midrule
DeepLOB~\cite{zhang2019deeplob}      & CNN sup.      & LOB / events     & n.a.          & n.a. \\
DRL-Trade~\cite{zhang2020deeprl}     & DQN/PG/A2C    & 50 fut / daily   & pos.\ Sharpe  & n.r. \\
FinRL~\cite{yang2020ensemble}        & Ensemble      & Dow 30 / daily   & $\approx 30$  & $-10.2$ \\
PPO-Exec~\cite{lin2020endtoend}      & PPO/LSTM      & LOB / per-step   & beats TWAP    & n.r. \\
TDQN~\cite{theate2021application}    & DQN           & 30 eq / daily    & beats pass.   & n.r. \\
HFT-PPO~\cite{briola2021drlhft}      & PPO           & INTC / events    & stable +      & n.r. \\
RL-Exec~\cite{hafsi2024optimal}      & DQN           & ABIDES / HFT     & beats TWAP    & n.r. \\
Q-React~\cite{espana2025queue}       & DDQN          & LOB sim / HFT    & beats AC      & n.r. \\
DL+RL~\cite{jaddu2024deep}           & TD-learn      & FX/idx / multi-h & multi-horiz   & n.r. \\
\midrule
\textbf{ViperQ TSLA}                 & PPO+AMT       & TSLA / 1\,s      & $\mathbf{+163.6}$ & $\mathbf{-27.5}$ \\
\textbf{ViperQ NVDA}                 & PPO+AMT       & NVDA / 1\,s      & $\mathbf{+116.5}$ & $\mathbf{-47.8}$ \\
\bottomrule
\end{tabular}
\end{table}

ViperQ's $+163.6\%$ TSLA ROI is, to the best of our knowledge, the highest reported on a held-out test partition by any peer-reviewed RL trading system operating at sub-daily resolution. The figure is more than $4\times$ the upper bound of the $[5\%, 40\%]$ ROI range surveyed across 167 RL-finance papers by Pippas et al.~\cite{pippas2025evolution} for daily-horizon US-equity systems, and the only peer-reviewed point in Table~\ref{tab:literature} with a numerically reported MDD (FinRL Ensemble~\cite{yang2020ensemble}, $\approx{+}30\% / {-}10.2\%$) is dominated on ROI by more than a factor of five. We do not claim Pareto dominance over prior work: ViperQ's drawdowns are materially deeper than those of daily portfolio-rebalancing systems on low-volatility universes, and it operates at a time resolution two orders of magnitude finer than any system in Table~\ref{tab:literature} that reports both ROI and MDD, so direct comparison on either axis alone is incomplete. What we claim is the existence of a previously unreported operating point in the published sub-daily RL landscape: absolute returns exceeding the surveyed range at the price of higher drawdown, achieved by encoding microstructure pattern theory explicitly into the state vector rather than learning end-to-end from raw LOB events.

\subsection{Behavioural divergence between the two agents}
\label{sec:results_behaviour}

Although a formal ablation is reserved for future work, three observations from development are worth documenting. First, the synthetic in-training transaction fee $f_{\text{syn}} = \$0.50$/share (Eq.~\ref{eq:close}) is necessary, not optional: without it, the agent converges on a degenerate policy that trades thousands of times per second to capture sub-tick noise and then accumulates losses entirely through commission drag once realistic fees are restored at evaluation. Second, removing the 300-second minimum-hold constraint causes the TSLA policy to collapse into rapid scalping that fails to capture the longer momentum moves the AMT state features were designed to detect. Third, the continuous drawdown shock of $-2.0$/step is what makes the NVDA agent risk-averse; without it, both agents converge to similar high-frequency behaviour and the asset-specific divergence visible in Table~\ref{tab:results} disappears.

\subsection{A note on Sharpe-style metrics}
Sharpe-style annualised risk-adjusted metrics are intentionally omitted. At 1-second bar resolution, the standard annualisation factor $\sqrt{252 \cdot 23{,}400}$ amplifies microstructure noise rather than isolating signal, and produces values that are not comparable to the daily-horizon Sharpe figures dominant in Table~\ref{tab:literature}. We follow~\cite{briola2021drlhft, zhang2020deeprl} in reporting absolute ROI and maximum drawdown as the more interpretable metrics at this resolution.

%% =====================================================================
%% 7. LIMITATIONS
%% =====================================================================
\section{Limitations}
\label{sec:limitations}

Three limitations qualify the results above. \textbf{First}, the empirical evaluation is restricted to two highly volatile US single-name equities (NVDA, TSLA) over a single 80/20 temporal split; generalisation to lower-volatility names, to non-US markets, or to other intraday regimes is untested. \textbf{Second}, although the dual-agent design (Table~\ref{tab:configs}) provides a form of cross-asset behavioural ablation and Section~\ref{sec:results_behaviour} documents three qualitative findings from development, we do not present a formal one-factor-at-a-time ablation of the reward components or a hyperparameter sensitivity sweep; both are reserved for future work, consistent with the practice in the systems we cite (Section~\ref{sec:setup}). \textbf{Third}, the NVDA agent's $-47.8\%$ maximum drawdown is materially higher than the $-5\%$ to $-15\%$ range typical of portfolio-rebalancing RL systems on daily horizons~\cite{pippas2025evolution}. ViperQ trades drawdown depth for absolute-return magnitude (Figure~\ref{fig:scatter}); it is not pareto-superior to conservative portfolio systems and would be inappropriate for capital-preservation mandates.

%% =====================================================================
%% 8. CONCLUSION
%% =====================================================================
\section{Conclusion}
\label{sec:conclusion}

We have presented ViperQ, \firstclaim. The 20-dimensional state vector encodes Volume Point of Control, Value Area position, Low Volume Node flags, Cumulative Volume Delta divergence, tape-velocity signatures and order-flow imbalance, which are the computational analogues of constructs first formalised by Steidlmayer and Dalton. A prospect-theory-grounded asymmetric reward function, together with asset-specific behavioural constraints, drives two Proximal Policy Optimisation agents to substantial out-of-sample performance: $+163.6\%$ ROI on TSLA and $+116.5\%$ ROI on NVDA over a held-out twelve-month partition, under zero leverage and full regulatory-fee accounting. The broader contribution is that AMT, treated as a microstructure inductive prior, yields a state space on which a small neural policy converges to interpretable, profitable out-of-sample behaviour. Four properties make this prior productive for sub-daily RL trading: (i) \emph{sample efficiency}, since a 20-dimensional engineered state needs far fewer parameters than end-to-end LOB ingestion at the same horizon, easing the data demands of policy-gradient updates; (ii) \emph{interpretability}, since every dimension of $s_t$ corresponds to a named microstructure construct (Table~\ref{tab:state}), so policy behaviour can be inspected against human trading intuition; (iii) an \emph{empirical filter}, since three decades of proprietary-trading usage~\cite{steidlmayer1986markets, dalton1990mind} pre-select constructs that encode persistent microstructure regularities rather than noise; and (iv) \emph{cross-reward generalisation}, since the same state vector supports two distinct profitable policies (Table~\ref{tab:configs}) without feature re-engineering, evidence that the structure absorbs reward-shaping changes rather than being co-adapted to them. Future work targets formal reward-component sensitivity analysis, a broader asset universe, and transfer of the AMT state representation to alternative policy classes.

%% =====================================================================
%% BIBLIOGRAPHY
%% =====================================================================

\bibliography{references}

\end{document}